\documentclass[letterpaper]{article} 
\usepackage{aaai23}

\usepackage{times}  
\usepackage{helvet}  
\usepackage{courier}  
\usepackage[hyphens]{url}  
\usepackage{graphicx} 
\usepackage{natbib}  
\usepackage{caption} 
\usepackage{algorithm}
\usepackage{algorithmic}
\usepackage{xcolor}
\usepackage{amsmath}

\usepackage{newfloat}
\usepackage{listings}
\DeclareCaptionStyle{ruled}{labelfont=normalfont,labelsep=colon,strut=off} 
\floatstyle{ruled}
\newfloat{listing}{tb}{lst}{}
\floatname{listing}{Listing}
\usepackage{subcaption}

\makeatletter
\def\thickhline{%
  \noalign{\ifnum0=`}\fi\hrule \@height \thickarrayrulewidth \futurelet
   \reserved@a\@xthickhline}
\def\@xthickhline{\ifx\reserved@a\thickhline
               \vskip\doublerulesep
               \vskip-\thickarrayrulewidth
             \fi
      \ifnum0=`{\fi}}
\makeatother

\newlength{\thickarrayrulewidth}
\usepackage{enumitem}
\usepackage{makecell}
\usepackage{comment}

\title{Action Dynamics Task Graphs for Learning Plannable Representations of Procedural Tasks}
\author{
    Weichao Mao,\textsuperscript{\rm 1,2}\thanks{Work done while interning at Reality Labs Research, Meta.}
    Ruta Desai,\textsuperscript{\rm 2}
    Michael Louis Iuzzolino,\textsuperscript{\rm 2}
    Nitin Kamra\textsuperscript{\rm 2}
}
\affiliations{
    \textsuperscript{\rm 1} University of Illinois Urbana-Champaign\\
    \textsuperscript{\rm 2} Reality Labs Research, Meta\\

    weichao2@illinois.edu,
    \{rutadesai, mliuzzolino, nitinkamra\}@meta.com
}

\begin{document}

\maketitle

\begin{abstract}
Given video demonstrations and paired narrations of an at-home procedural task such as changing a tire, we present an approach to extract the underlying \emph{task structure} -- relevant actions and their temporal dependencies -- via action-centric task graphs. Learnt structured representations from our method, \emph{Action Dynamics Task Graphs (ADTG)}, can then be used for understanding such tasks in unseen videos of humans performing them. Furthermore, ADTG can enable providing user-centric guidance to humans in these tasks, either for performing them better or for learning new tasks. Specifically, we show how ADTG can be used for: (1) tracking an ongoing task, (2) recommending next actions, and (3) planning a sequence of actions to accomplish a procedural task. We compare against state-of-the-art Neural Task Graph method and demonstrate substantial gains on 18 procedural tasks from the CrossTask dataset, including 30.1\% improvement in task tracking accuracy and 20.3\% accuracy gain in next action prediction.
\end{abstract}

\section{Introduction}
With the advent of augmented reality and advanced vision-powered AI systems, we envision a future of next generation AI assistants that will be able to deeply understand the at-home tasks that users are doing from visual data and assist them to accomplish these tasks. These AI assistants with reasoning capabilities would be able to track the user's actions in an ongoing complex task, detect mistakes, and provide actionable guidance to the users such as next steps to take. Such user-centric guidance can either help the user better perform a task or help them learn a new task more efficiently.

To make progress toward such assistants, we focus on at-home \emph{procedural tasks}, where humans routinely require guidance. Examples of such tasks include assembling furniture, making lemonade, preparing fish curry, changing a tire, and more. Procedural tasks typically involve executing specific durative actions in certain temporal order. We refer to the actions needed to accomplish a given procedural task and their temporal dependencies as \emph{task structure}. Our goal is to learn representations that capture such underlying task structure for downstream guidance generation from visual demonstrations such as videos and annotated action labels.

Learning such task structure from videos is challenging for multiple reasons. First, procedural tasks require a representation to track the state of the task and identify durative actions grounded in visual observations. Both these challenges require dealing with immense variations in visual observations even for demonstrations from a single task. Also, procedural tasks often have multiple acceptable action sequences; i.e., the ordering of some actions may be interchanged without affecting the final outcome while certain actions are temporally dependent on others. Such interchangeability and temporal dependence between actions must also be learnt directly from data.

To mitigate these challenges, we propose a structured representation -- Action Dynamics Task Graph (ADTG), a graph data-structure centered around actions that inherently captures the temporal dependence in procedural tasks. ADTG focuses solely on actions and avoids representing states in the graph, thereby making the size of the graph much smaller than typical task graph representations. It uses robust visual representations of actions learnt by treating actions as ``transformations between states''. We also present an approach to learn: (i) task tracking and (ii) next action prediction models based on ADTG using video demonstrations and paired action annotations of a procedural task.

Our approach allows us to observe users while they perform procedural tasks and generate actionable plans for them from direct visual observations. Specifically, ADTG can enable:~(1)~tracking of an ongoing task using the learnt graph structure,~(2)~recommendation of a next step to the user, and~(3)~planning a sequence of potential next steps to complete the task. We compare our method against the Neural Task Graph (NTG) method~\cite{huang2019neural} and demonstrate substantial performance gains on the CrossTask dataset~\cite{zhukov2019cross}. Specifically, our method achieves 30.1\% improvement in task tracking accuracy and 20.3\% improvement in next action prediction accuracy. We also present an analysis of its plan generation capability, which is not possible with NTG, and show further ablation studies to understand its strengths and weaknesses.

\section{Related Work}

Existing work in learning to plan from video demonstrations can be broadly categorized into implicit and explicit approaches based on whether it explicitly maintains structural representations of the tasks.

\subsubsection{Implicit Representations} Many existing approaches learn to plan directly from visual observations without explicitly characterizing the underlying structure of the task~\citep{srinivas2018universal,zhukov2019cross,sun2022plate,zhao2022p3iv}. In particular, under the assumption of a differentiable action space, Universal Planning Networks (UPN)~\citep{srinivas2018universal} use gradient descent to directly learn the planner and its representations in an end-to-end fashion, by optimizing a supervised imitation learning objective. \citet{kurutach2018learning} combines representation learning and planning using an InfoGAN to learn a generative model of sequential observations and a low-dimensional planning model. More recently, a Transformer-based planning network named PlaTe~\citep{sun2022plate} has been proposed for procedure planning in instructional videos, which simultaneously learns the planning model and the latent semantic representations.
In comparison, our approach tries to explicitly learn structural representations of tasks, leading to a modular pipeline that allows us to easily achieve various downstream learning objectives such as task tracking, action recommendation, and planning. 

\subsubsection{Explicit State-Centric Representations} Another line of work uses more explicit representations for planning. These are often the state transition model and the policy from a Markov Decision Process (MDP), and are learnt using a combination of model-based deep reinforcement learning and imitation learning approaches directly from user demonstrations~\citep{fang2020dynamics,chang2020procedure,bi2021procedure}. In case of visual observations, one often resorts to learning a partially observable MDP structure~\citep{hafner2019learning}. Other approaches try to represent state transitions via transition graphs~\citep{liu2016jointly} and learn them from annotated data~\citep{pan2020multi,xu2020benchmark}. However, directly working on the high-dimensional visual observations often leads to graphs with a prohibitively large number of states, which are computationally intractable for planning.  

\subsubsection{Explicit Action-Centric Representations} To avoid the intractability of state-centric representations, some recent works have focused on action-centeric representations because the action space of a task is typically much smaller than its observation/state space. The key idea is to leverage the Conjugate Task Graph (CTG) initially proposed by~\citep{hayes2016autonomously}, which reverses the roles of states and actions in the task representations. More recent approaches have proposed other easier to learn variants of CTGs~\citep{huang2019neural,chang2020procedure}, but often do not support flexible multi-step planning. Our approach is centered around building a variant of the CTG that abstracts out states and learns the inter-dependence amongst actions, with a key focus on flexible multi-step plan generation at runtime.

\section{Problem Formulation}

In this section, we introduce procedural tasks and graph-based representations used in the recent literature to capture their structure. We also briefly describe the CrossTask dataset used in our experiments.

\subsection{Procedural Tasks}
\label{subsec:proc_tasks}

\textbf{Definition  } Let the assistive agent be located in a world with an underlying state $s$, which also includes information about the user being assisted. Let $g$ be a set of desired goal states. The agent can suggest durative actions from the set $A=\{a_1, \ldots, a_N\}$ to the user. Each action $a_i$ has pre-conditions which must be met by the current state $s$, before the action can be enacted. Further, each action has effects (a.k.a. post-conditions) which change the current state $s$, thereby transitioning the world into a new state $s'$. Accomplishing a \emph{procedural task} requires finding a sequence of actions from the current world state $s$ to achieve a desired goal $g$. Many at-home tasks like: make lemonade, change a tire, etc. can be captured by this formulation.

\noindent
\textbf{Assumptions  } In our work, we assume the set of durative actions $A$ for the procedural task is known a priori. The trained agent should observe a user perform a procedural task and recommend a relevant next action towards a desired goal state for the task. At test time, we only have access to input video frames\footnote{optionally, also audio} of the user performing the task. At training time, we additionally also have action labels annotated on the input videos along with their temporal start and end boundaries. We do not have direct access to the current state $s$ at any time, neither the pre-conditions which enable actions, nor the post-conditions of actions. Hence, our agent must represent the current task state, learn to track the state, detect the feasible next actions, plan to achieve a goal state and recommend the next action in the plan to the user.

\noindent
\textbf{Challenges  } This is challenging for three reasons. The first challenge this introduces is to define and learn a state representation in order to track the state of a procedural task. The second challenge is to ground the action set $A$ in visual observations. Both these challenges require dealing with immense variations in visual input even for demonstrations from a single task. Finally, there are often multiple acceptable action sequences for procedural tasks. The ordering of some actions may be interchanged without affecting the final outcome. For instance, in the task ``Make lemonade'', the actions ``pour lemon juice'' and ``pour water'' may be interchanged. However, not all sequences are valid since there exist temporal dependencies between certain actions, e.g., the action ``squeeze lemon'' must occur before ``pour lemon juice'' because the former's effects are pre-conditions for the latter. Hence, the third challenge is to learn the interchangeability and temporal dependence between actions from demonstrations of the task.

\subsection{Graph-based Task Representations}

\begin{figure*}[!hbt]
\centering
\begin{subfigure}{.45\textwidth}
    \includegraphics[width=\textwidth]{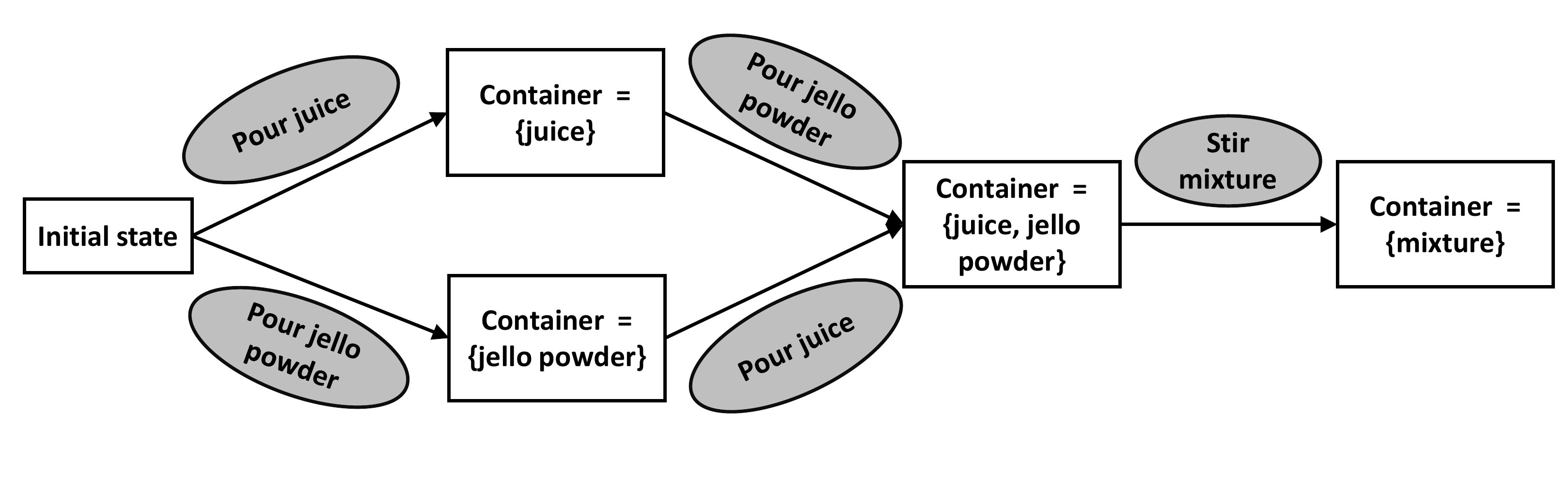}
    \caption{Task Graph}
    \label{subfig:TG}
\end{subfigure}
\hfill
\begin{subfigure}{.45\textwidth}
    \includegraphics[width=\textwidth]{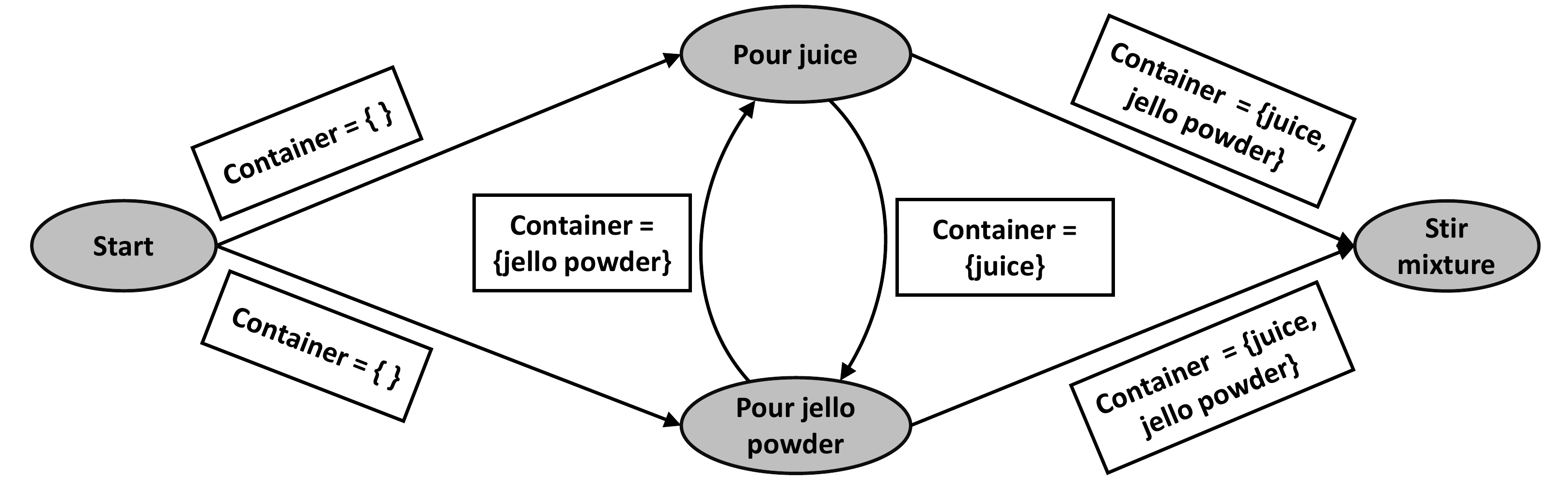}
    \caption{Conjugate Task Graph}
    \label{subfig:CTG}
\end{subfigure}
\caption{Comparing graph-based task representations for a subset of actions on the ``Make jello shots'' task. }
\label{fig:tg-ctg}
\end{figure*}

Many approaches in existing literature~\cite{chang2020procedure,bi2021procedure} have traditionally used \emph{Task Graphs} to model states, actions, and the state transitions for procedural tasks. Task Graphs represent states on their nodes and actions on their edges to represent transitions between states caused due to actions (Figure~\ref{fig:tg-ctg}(a)). However, due to immense variation in visual observations, observation spaces (and hence state spaces) can be extremely large. Hence, Task Graphs can have a prohibitively large set of nodes, making them intractable for planning purposes~\cite{huang2019neural}.

The above challenges can be mitigated by focusing on two key observations: (a) action spaces of tasks are much smaller than their visual observation and state spaces, and (b) procedural tasks primarily require grounding actions in observations and modeling their inter-dependence for task completion. These observations suggest using a data structure centered around actions for modeling procedural tasks. Hence, more recent methods focus on (variants of) \emph{Conjugate Task Graphs} (CTGs), which leverage the conjugate relationship between states and actions by reversing their representations in the Task Graph~\citep{hayes2016autonomously}.

A CTG represents actions on its nodes, while a directed edge from action $a_i$ to $a_j$ captures the pre-conditions of action $a_j$ met by $a_i$. Hence, a CTG has far fewer nodes than a Task Graph and only captures the part of state space relevant to the actions involved in the task, while ignoring other nuisance factors in the potentially infinite state space. Figure~\ref{fig:tg-ctg} illustrates a Task Graph and a CTG on a subset of actions for the ``Make jello shots'' task. The original design of CTGs still tries to encode state information into the edges~\citep{hayes2016autonomously}. However, directly learning state representations from video demonstrations is often hard without extensive object and attribute labeling. Since such annotations are costly to procure, recent approaches~\cite{huang2019neural} often learn simplified versions of CTGs which abstract out state information. In our approach, we shall adopt a similar approach to learn a simplified variant of CTG.

\subsection{CrossTask dataset}

We chose the CrossTask dataset~\citep{zhukov2019cross} to learn and evaluate our models on real-world instructional videos. This dataset contains 2750 videos, each demonstrating one of its 18 procedural tasks (e.g., ``Make a latte'') and these demonstrations span a variety of visual variations as well as executed action orderings. In Appendix~\ref{app:dataset}, we provide a detailed description of the CrossTask dataset and more statistics for the 18 tasks.

For reproducibility and fair comparison with existing methods, we leverage the pre-computed 3200-dimensional video features provided along with the CrossTask dataset for every one-second segment of the videos. Manually annotated action labels and their corresponding temporal segmentation boundaries are also provided for learning and evaluation. Following~\citet{zhukov2019cross}, for each task we use 50 videos for training, 20 videos for validation of hyperparameters, and leave the rest of the videos unseen for testing.

\section{Action Dynamics Task Graphs}

\begin{figure*}[!t]
\centering
\begin{subfigure}{.32\textwidth}
    \includegraphics[width=\textwidth]{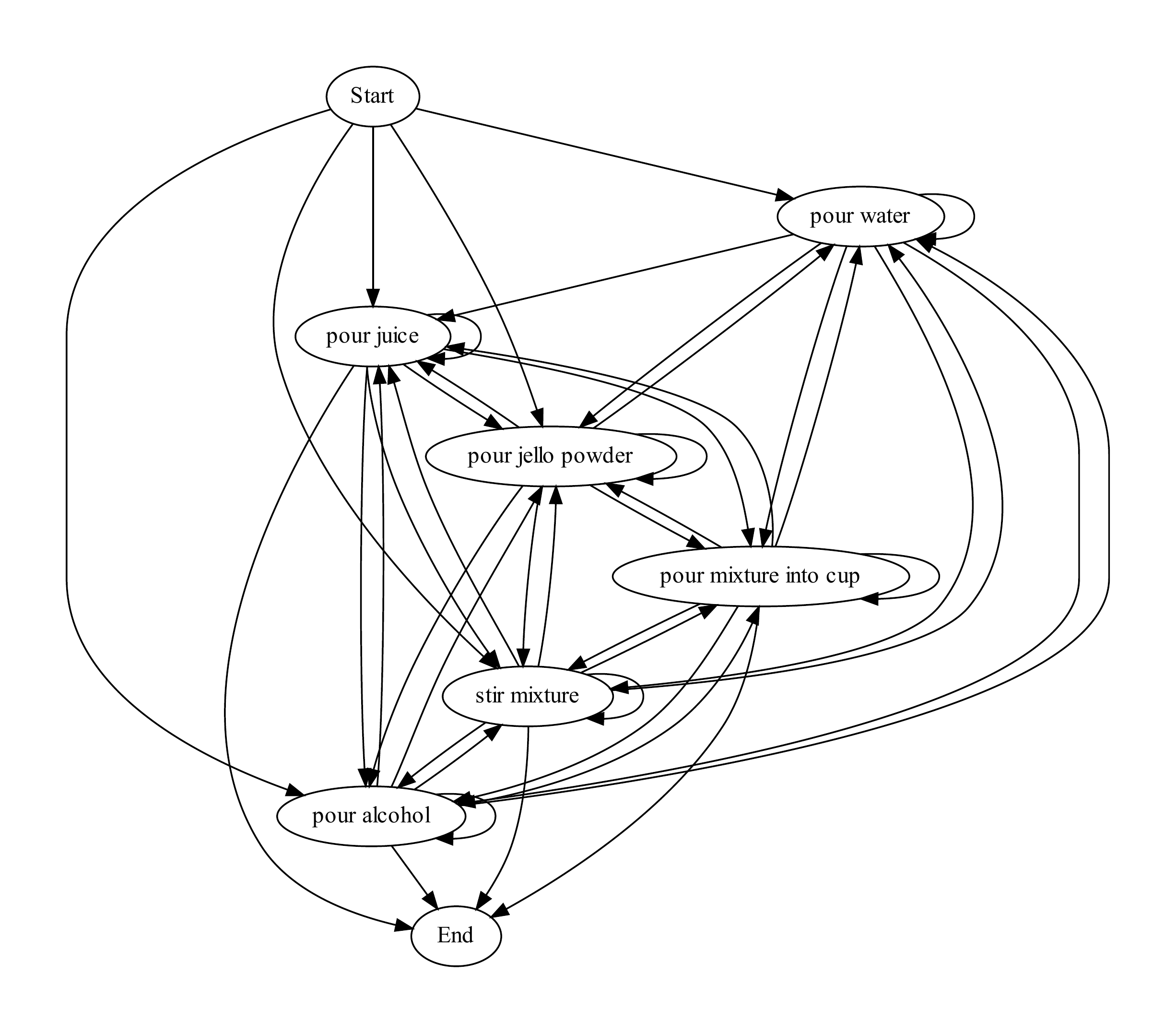}
    \caption{``Make jello shots''}
\end{subfigure}
\hfill
\begin{subfigure}{.32\textwidth}
    \includegraphics[width=\textwidth]{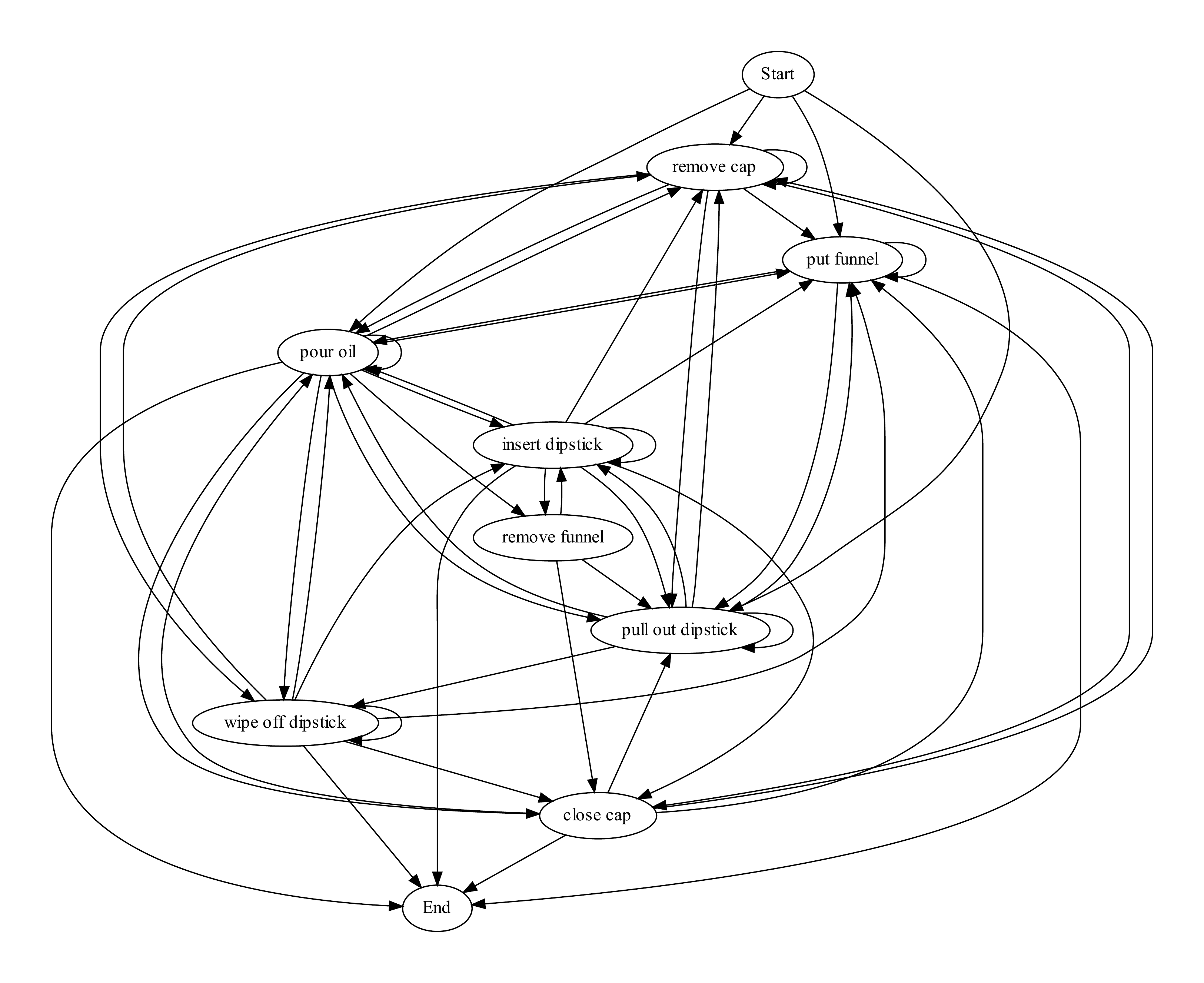}
    \caption{``Add oil to your car''}
\end{subfigure}
\hfill
\begin{subfigure}{.32\textwidth}
    \includegraphics[width=\textwidth]{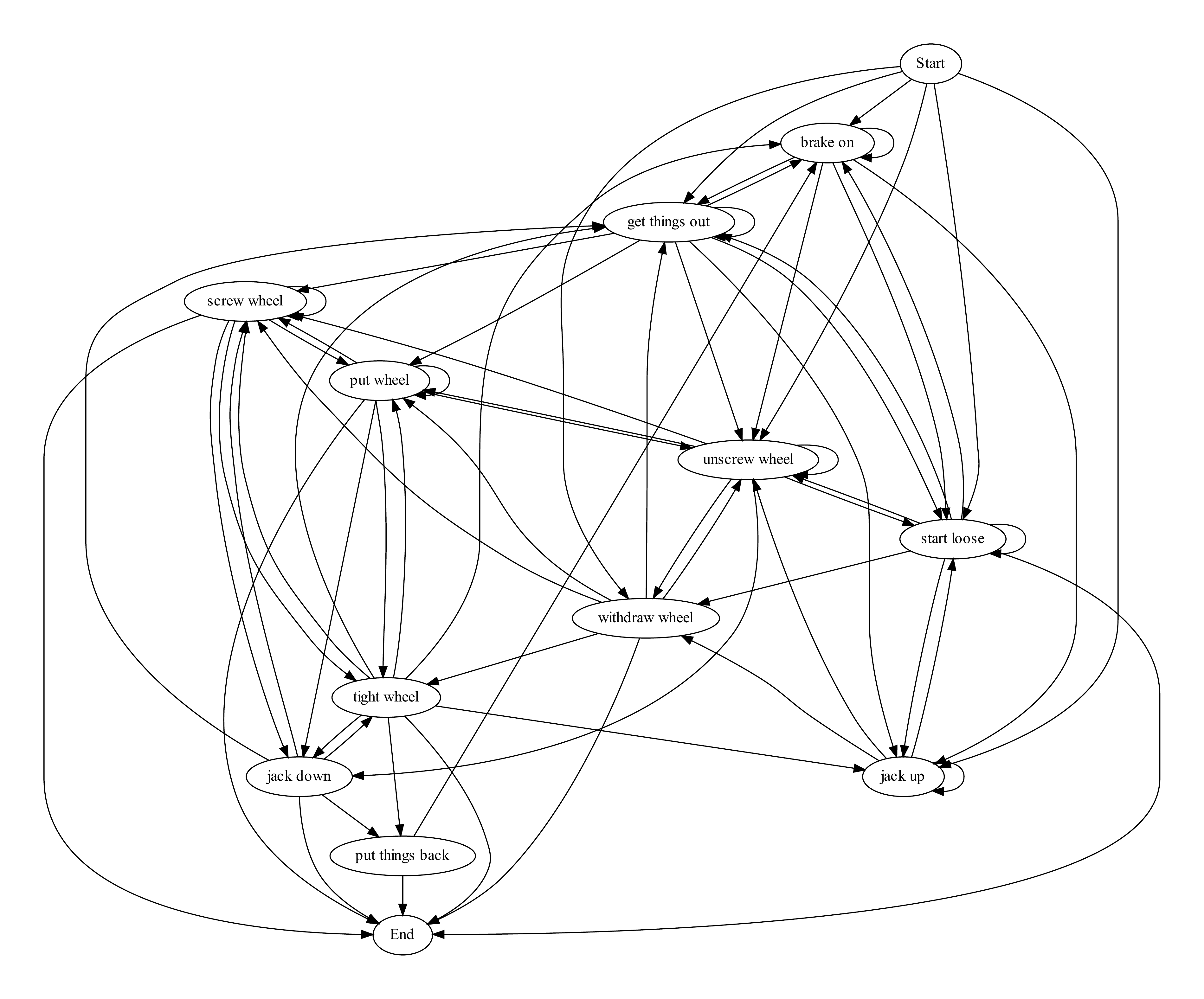}
    \caption{``Change a tire''}
\end{subfigure}
\caption{Action Dynamic Task Graphs for three tasks from the CrossTask dataset. }
\label{fig:example-ctgs-app}
\end{figure*}

In this section, we present our approach for learning plannable representations and using them to provide guidance for procedural tasks. We choose to represent a procedural task with a simplified version of a CTG, which we call as an Action Dynamics Task Graph (ADTG)\footnote{Please excuse the notational abuse, where we refer to both our overall approach as well our CTG variant as Action Dynamics Task Graph (ADTG). The one being referred to will be generally clear from context.}.
Our overall approach contains several trained modules: In Section~\ref{subsec:generating_adtg}, we introduce how we generate the underlying graphs and the associated action embeddings from demonstrations. In Section~\ref{subsec:adtg_guidance}, we discuss how the generated graphs are used to provide guidance to the human users in multiple scenarios, including task tracking, next action prediction, and plan generation. Implementation details are discussed in Section~\ref{subsec:implementation}.

\subsection{Generating ADTG}\label{subsec:generating_adtg}
While a CTG stores the effects of actions on its edges, it is generally hard to obtain this information from annotations since a universal grammar for object definitions and their attributes must be established and taught to the annotators. Consequently, state-specific information is generally absent from most existing procedural task datasets. However, actions are generally known a priori, form a smaller set and are relatively easier to annotate in videos. Hence, the graph variant we devise (namely, ADTG) abstracts the notion of state completely and focuses solely on actions and identifying their inter-dependence. However, the state of the task can still be tracked using the graph itself.

Each node in an ADTG represents an action $a_i$. A directed edge from action $a_i$ to $a_j$ encodes a temporal dependence from $a_i$ to $a_j$, if the latter action has been observed to occur directly following the former during any demonstration of the task. To generate the ADTG graph structure of a single task $T$ from all associated demonstrations in the training set, we perform the following steps:
\begin{enumerate}
    \item Extract all actions that appeared in any video corresponding to the task $T$ into an action set $A_T$ for the task.
    \item Initialize the ADTG graph with one node for each action in the set $A_T$.
    \item Iterate over all videos for the task $T$:
    \begin{enumerate}
        \item Let $(a_1, a_2, \ldots, a_n)$ be an action sequence encountered in a single video.
        \item For every pair of consecutive actions $(a_t,a_{t+1})$ in the sequence, add a directed edge from the action node of $a_t$ to that of $a_{t+1}$ in the ADTG (if it does not exist already).
    \end{enumerate}
\end{enumerate}
By construction, if both edges $a_i \rightarrow a_j$ and $a_j \rightarrow a_i$ exist in an ADTG, it makes the actions' ordering independent of each other. On the other hand, if only the edge $a_i \rightarrow a_j$ exists in the graph, it encodes a temporal constraint between the two nodes, potentially because the action $a_i$ could be contributing some pre-conditions for the action $a_j$. In Figure~\ref{fig:example-ctgs-app}, we illustrate the ADTGs for several tasks from the CrossTask dataset.

\begin{figure*}[!t]
\centering
\begin{minipage}{.55\textwidth}
  \centering
  \includegraphics[width=\linewidth]{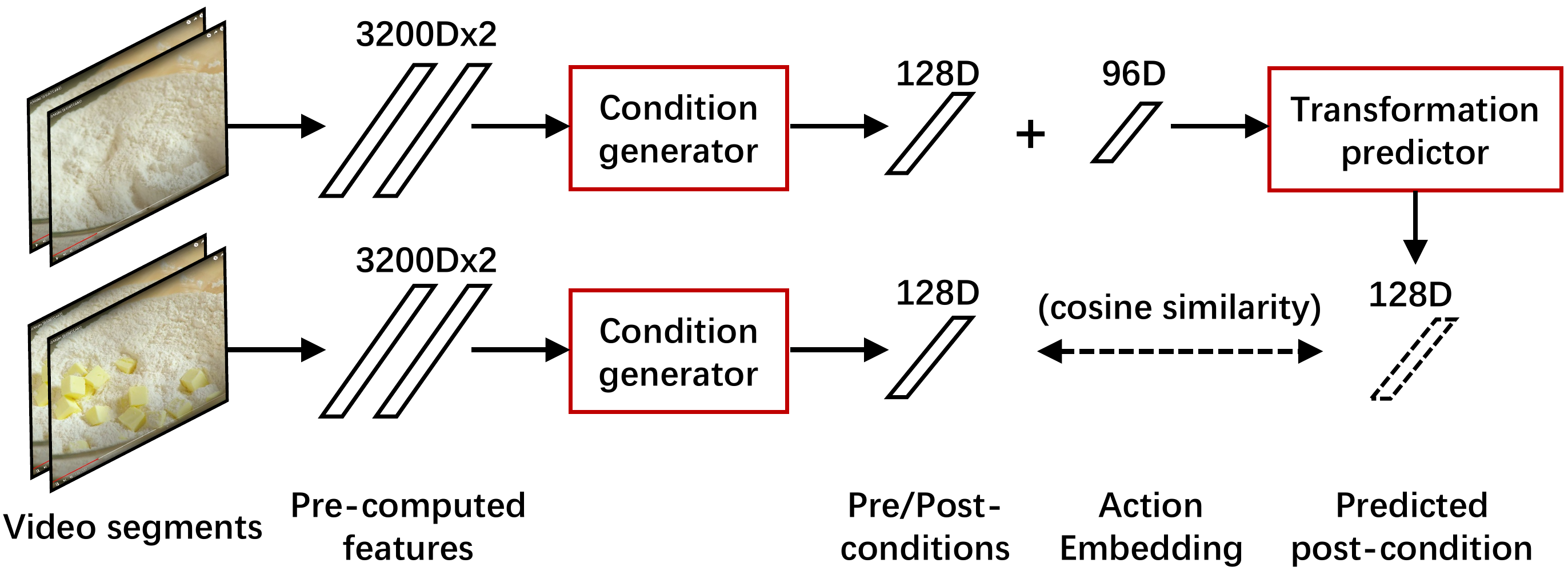}
  \captionof{figure}{Action embedding network}
  \label{fig:action_embedding}
\end{minipage}%
\hfill
\begin{minipage}{.4\textwidth}
  \centering
  \includegraphics[width=\linewidth]{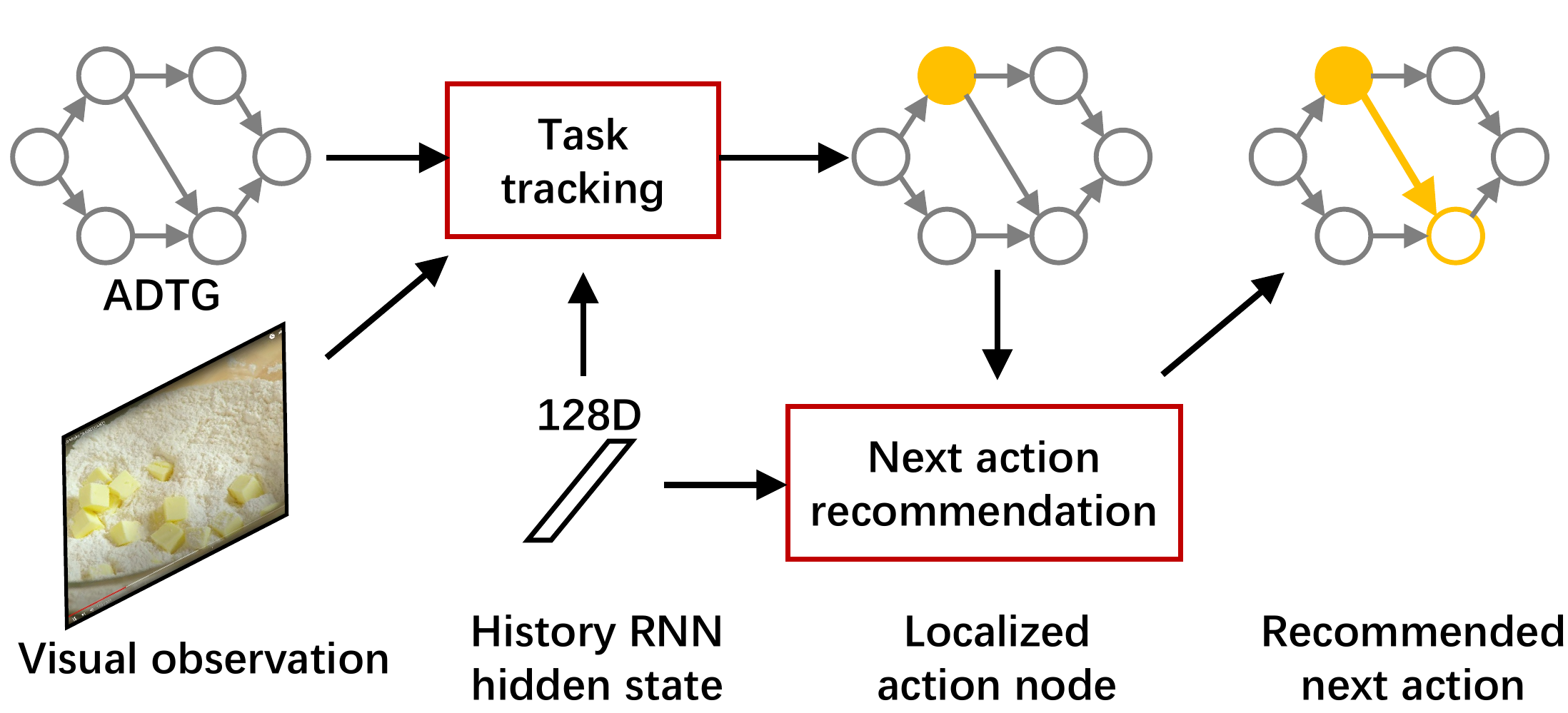}
  \captionof{figure}{Using ADTG for guidance}
  \label{fig:using_adtg}
\end{minipage}
\end{figure*}

\subsubsection{Action Embedding Network}
The ADTG method also associates each action with an embedding vector representation in a continuous space for downstream use. To learn the action embeddings, we leverage the fact that actions are transformations from their pre-conditions to their effects (inspired from~\citep{wang2016actions}). Illustrated in Figure~\ref{fig:action_embedding}, the \emph{action embedding network} consists of three learnable components: a condition generator, a condition transformation prediction and the action embeddings.

The condition generator transforms the pre-conditions and post-conditions from the visual feature space to lower-dimensional semantic feature vectors. We use two-second video segments around the beginning and ending of an action to generate the pre- and post-condition semantic features respectively. Specifically, a video from the CrossTask dataset can be represented as $X=(x_1, x_2, \dots,x_T)$, where $T$ is the time duration of the video, and $x_t$ is the 3200-dimensional feature vector encoding the $t$-th second of the video. During training, the segmentation boundaries of the actions are given in the form of time intervals. For any action $a$, suppose that it occurs during the time interval $[t_1,t_2]$ (rounded to nearest integers). We use the two visual feature vectors $X_{\text{pre}} = (x_{t_1-1}, x_{t_1})$ to generate the pre-condition semantic features $f(X_{\text{pre}})$ for the action, and $X_{\text{post}} = (x_{t_2}, x_{t_2+1})$ to generate the post-condition semantic features $f(X_{\text{pre}})$. The condition generator is shared over all actions, for generating both pre- and post-conditions.

Each action $a$ is associated with a trainable action embedding vector $e_a$. We next introduce a \emph{transformation predictor} $g$ to train the action embeddings so as to capture the transformation from the pre-conditions to post-conditions. The transformation predictor takes as input a pre-condition $f(X_{\text{pre}})$ and a candidate action embedding $e_{a}$, and outputs a predicted post-condition $g(f(X_{\text{pre}}), e_{a})$. If the action $a$ matches the pre-condition, we would expect the predicted post-condition to be close to the ground-truth post-condition. We define the distance between two post-condition vectors $v_1$ and $v_2$ as: $D(v_1, v_2) = 1-\frac{v_1 \cdot v_2}{\|v_1\| \|v_2\|}. $ We train the aforementioned modules (condition generator, action embeddings, and transformation predictor) together to minimize the distance for the matching action $a$, as captured by the following discriminative loss term:
\begin{equation}\label{eqn:loss1}
    \mathcal{L}_{disc}= D(g(f(X_{\text{pre}}), e_{a}), f(X_{\text{post}})).
\end{equation}
To avoid trivial solutions, we further add a contrastive loss term to maximize the distance for incorrect actions $a' \neq a$:
\begin{equation}\label{eqn:loss2}
    \mathcal{L}_{cont} = \sum_{a'\neq a} \max(0, M-D(g(f(X_{\text{pre}}), e_{a'}), f(X_{\text{post}}))),
\end{equation}
where $M$ is a margin threshold in the sense that we will not penalize an incorrect action if its distance is already larger than $M$. Combining Equations~\eqref{eqn:loss1} and~\eqref{eqn:loss2} gives the complete loss term $\mathcal{L} =  \mathcal{L}_{disc} +  \mathcal{L}_{cont}$ for training the action embeddings. We train the aforementioned modules using all the actions jointly from the training videos for all the tasks and store the ADTG graphs and the action embedding vectors $\{e_a\}_{a \in A}$ as representations for the actions.

\subsection{Using ADTG for Guidance}\label{subsec:adtg_guidance}
We now introduce how our approach utilizes the ADTG representations to provide guidance to humans when they perform a procedural task. We consider multiple scenarios of guidance, including task tracking, next action recommendation, and plan generation. A working pipeline of using ADTG is illustrated in Figure~\ref{fig:using_adtg}. 

\subsubsection{Task Tracking and Next Action Recommendation}
In \emph{task tracking}, our model tries to track the human's progress when performing a certain task, by localizing the current visual observation in the corresponding ADTG graph. 
For any time step $t\in\{1,\dots, T\}$ of a given video, the inputs to the task tracking module include the current visual observation $x_t$, the action history information $h_{t-1}$ up to time step $t-1$, and the embedding of a candidate action $e_a$ being evaluated. The task tracking module then outputs a confidence score indicating whether the candidate action $a$ captures the given visual observation $x_t$. We then enumerate all the candidate actions and localize to the action node $a_t$ with the highest confidence score in the ADTG. In particular, to get a succinct representation of the action history, we pass the sequence of history action embeddings $(e_{a_1}, \dots, e_{a_{t-1}})$ up to time $t-1$ through a recurrent neural network (RNN), and use the RNN hidden state as the action history information $h_{t-1}$. We train the task tracking module and the history RNN over all the time steps of the training videos, and use the cross entropy loss to update the network parameters for both of them using back-propagation. 

In \emph{next action recommendation}, our model recommends to the user an action to perform in the next time step. For any time step $t$, the next action recommendation module takes as input the currently localized action node embedding $e_{a_t}$, the action history information $h_{t}$, and the embedding of a candidate next action $e_a$ being evaluated. It maps these inputs to a confidence score as the output, indicating where it believes that the evaluated candidate action is a suitable one for the next time step. We then enumerate all the candidate next actions from the outgoing edges of the localized action node in the ADTG, and select the action with the highest confidence score as the recommendation $a_{t+1}$ to the user. The action history information comes from the hidden state of the same history RNN as in the task tracking module. We again use the cross entropy loss to train the next action recommendation module over any consecutive action pairs $(a_t, a_{t+1})$ in the training videos. 

\subsubsection{Plan Generation}
Given a visual observation, \emph{plan generation} is used to generate a sequence of actions to accomplish the task from this point onward. The visual observation can come from either the very beginning of the task (i.e., complete plan generation) or from a random midpoint in the execution of the task (i.e., planning after a prefix observation sequence). In our approach, plan generation is done by applying one step of task tracking, followed by multiple steps of next action recommendation in an autoregressive way. Specifically, given an initial visual observation and initial action history, we pass them through the task tracking module to recognize the current action, and localize to an action node in the ADTG. Using the embedding of the localized action and the action history information as input, we obtain the next action to perform from the next action recommendation module. Finally, we update the action history information by passing the newest action through the history RNN, and recursively invoke the next action recommendation module to generate a sequence of actions to perform. Repeating this process will lead to a plan to accomplish the task. In this sense, our plan generation is built entirely upon existing modules. 

In the recursive process of plan generation, instead of following the greedy action choice outputted by the next action recommendation module at each step, we use beam search to select the output sequence with higher overall likelihood. Specifically, for a beam search width of $k$, we maintain $k$ candidate action trajectories, perform rollouts based on these trajectories, and only keep the rollout trajectories with the top $k$ highest likelihoods at each step. In this way, our method strikes a good balance between computation complexity and the optimality of the selected action sequence.

\subsection{Implementation Details}\label{subsec:implementation}
In our implementation, the condition generator, transformation predictor, task tracking module, and the next action recommendation module are all instantiated as two-layer feed-forward neural networks. The pre- and post-conditions are $128$-dimensional semantic vectors, and the action embedding vectors are $96$-dimensional. The hidden size of the history RNN is $128$. The margin threshold for contrastive loss in Equation~\eqref{eqn:loss2} is set to be $M = 0.5$. In the plan generation module, the beam search width is $k=5$. We use the ADAM optimizer~\citep{kingma2014adam} with a learning rate of $1e-5$ for training the action embeddings, and learning rate $5e-5$ for training the task tracking and next action recommendation modules. The action embeddings and task tracking module are trained for $50$ epochs, while the next action recommendation modules takes $100$ training epochs. 

\section{Results}\label{sec:experiments}

In this section, we present the experimental results of our ADTG approach and compare with existing baselines. We evaluate ADTG on three important tasks that we have discussed: (1) task tracking, (2) next action recommendation, and (3) plan generation, on the 18 primary tasks of the CrossTask dataset.

\subsection{Comparison Baselines}
We compare our approach with three baselines from existing works, namely the Neural Task Graph (NTG) approach~\citep{huang2019neural}, the CrossTask (CT) approach~\citep{zhukov2019cross}, and a supervised-learning variant of the CrossTask approach (CT-S).
    
\subsubsection{NTG} Similar to ours, NTG~\citep{huang2019neural} is also a modularized method that uses (a simplified variant of) the conjugate task graphs as intermediate representations. While it supports task tracking and next action recommendation, its next action predictor requires visual observations as input. Hence, in the absence of an interactive environment, it cannot generate a plan for a task in an autoregressive way as we do. More details on NTG can be found in Appendix~\ref{app:ntg}.
    
\subsubsection{CT} The CrossTask baseline (see Appendix~\ref{app:crosstask}) refers to the weakly-supervised learning approach proposed in the same work~\citep{zhukov2019cross} along with the CrossTask dataset. It proposes a component model capable of performing task tracking, but does not support next action recommendation or plan generation. 

\subsubsection{CT-S} The original CrossTask approach is a weakly-supervised learning method. For fair comparison, we also consider a supervised-learning variant of CrossTask that adopts the same linear classifier as in~\citet{zhukov2019cross}, but further uses the annotated action segmentation boundaries during training.

\subsection{Evaluation Metrics}

We consider four metrics for evaluation, namely accuracy, accuracy excluding null actions, log-likelihood, and mean Intersection over Union
(mIoU).
\begin{itemize}[leftmargin=*]
    \item \emph{Accuracy}: It measures whether the action prediction of the model matches the ground-truth action at each individual step, and then averages over all time steps. 
    \item \emph{Accuracy excluding null actions}: The motivation for considering such a metric is that in the CrossTask dataset, about 72\% of the video duration does not have an actual action happening. These video segments contain introduction parts of the video or certain transitions from one action to another. We call these steps the \emph{null action} steps to distinguish them from the other steps that do have concrete actions. In task tracking, since we care more about the model's capability to correctly recognizing concrete actions, we introduce the ``accuracy excluding null actions'' metric to specifically measure the accuracy on the steps excluding the null actions. 
    \item \emph{Log-likelihood}: We compare the log-likelihood of the model prediction vs. that of the ground-truth action. If these two values are close enough, we would know that the model is not entirely wrong because, according to its predictions, it also has a high probability of selecting the ground-truth action as the output.
    \item \emph{Mean Intersection over Union (mIoU)}. This metric is specifically used for evaluating the plan generation performances. Let $\{a_t^*\}_{t\geq 1}$ be  the set of ground truth actions of a video, and $\{a_t\}_{t\geq 1}$ be the set of predicted actions. The mIoU is defined as $\frac{|\{a_t\}\cap \{a_t^\star\}|}{|\{a_t\}\cup \{a_t^\star\}|}$. Intuitively, to get a high mIoU score, the model needs to understand what actions are required, but need not be able to discern the correct ordering. It is hence a less strict metric than accuracy. 
\end{itemize}

\subsection{Experimental Results}

\subsubsection{Task Tracking}
Table~\ref{tab:task_tracking_results} shows the results for task tracking, which measures our method's capability of recognizing the action from a given visual observation. All results are averaged over 5 runs, and we show both the mean values and the standard deviations of the metric scores. The supervised variant of CrossTask achieves the highest overall accuracy. However, since about 72\% of the video duration are null action steps, the sample labels are largely biased toward null actions. Achieving a high overall accuracy does not necessarily indicate that the method is capable of correctly recognizing the actual actions, which is the main objective of task tracking. Instead, our ADTG approach achieves more than $30.0\%$ higher accuracy than the comparison baselines when excluding the null action steps, and at the same time shows a near-optimal overall accuracy. This suggests that ADTG is able to successfully track the task progress for most time steps of a given video and its predictions are not misled by the dominance of the null actions. We attribute the improvement to the incorporation of the action history information and the effective representation of the action embeddings. 

\begin{table}[!t]
\centering
\caption{Task tracking results}
\label{tab:task_tracking_results}
\begin{tabular}{ccccc}
\thickhline
Metrics & ADTG & NTG & CT & CT-S \\ \hline
Accuracy & \makecell{0.741 \\\small($\pm$ 0.013)} &  \makecell{ 0.663  \\ \small($\pm$ 0.014) }  & N/A & \makecell{\textbf{0.763}\\\small($\pm$0.005)} \\
\hline \makecell{Accuracy \\excl. null} & \makecell{\textbf{0.557} \\\small($\pm$ 0.042)} & \makecell{0.256 \\\small($\pm$ 0.025)} & \makecell{0.222\\\small($\pm$0.004)} & \makecell{0.224\\\small($\pm$0.004)} \\ \thickhline
\end{tabular}
\end{table}

\subsubsection{Next Step Recommendation}
Table~\ref{tab:next_action_results} shows the results for next action recommendation, which measures whether a method is capable of recommending a valid action to perform in the next time step. The candidate next action set consists of actions that have been observed to occur immediately after the current action in at least one training video demonstration. Our approach again achieves a higher accuracy than the NTG baseline, indicating that ADTG is able to recommend the ground-truth next action more frequently. Further, for ADTG, the difference between the average log-likelihood of the model predictions and that of the ground-truth actions is also smaller. This is especially important since often there are multiple possible correct next actions and the ground truth action in the dataset is only one of those. The smaller difference suggests that, though ADTG recommends an action different from the ground truth sometimes, the ground-truth actions also have a high probability to be selected. 

\begin{table}[!t]
\centering
\caption{Next action recommendation results}
\label{tab:next_action_results}
\begin{tabular}{ccc}
\thickhline
Metrics & ADTG & NTG \\ \hline
Accuracy & \makecell{\textbf{0.523}\\ {(\small $\pm$ 0.026)} } & \makecell{0.32 \\(\small$\pm$ 0.026)} \\
\hline Log-likelihood & \makecell{ $-$0.562\\(\small$\pm$0.015)} & \makecell{$-$ 0.018 \\(\small$\pm$ 0.001)}   \\
\hline \makecell{Log-likelihood\\ ground-truth} &  \makecell{$-$0.918\\(\small$\pm$0.023)} &  \makecell{$-$ 1.155 \\(\small$\pm$ 0.091)} \\ \thickhline
\end{tabular}
\end{table}

\subsubsection{Planning} The results for plan generation are shown in Table~\ref{tab:full_plan_results}. Neither NTG nor CT supports plan generation in the absence of an interactive environment, and hence only the results for ADTG are presented. We consider two cases here: In the \emph{complete plan generation} case, ADTG is used to generate the entire action sequence for a task from the very first step. The second case, namely \emph{planning after a prefix}, occurs when ADTG observes a human user performing the task up to a certain step, and then plans from this step onward. For this case, we uniformly sample a time step of the video, and use the ground-truth action sequence before this step as a prefix action history to generate the remaining actions.
From Table~\ref{tab:full_plan_results}, we see that ADTG achieves higher accuracy and mIoU scores in the planning after a prefix case. This is because ADTG uses an autoregressive beam search to plan, and errors are more likely to be accumulated for long horizon plans, as in the complete plan generation case. Finally, we remark that ADTG explicitly outputs an end-of-sequence (EOS) token to stop the recursive action generation process and terminate the planning. This allows ADTG to plan for a flexible horizon as opposed to existing methods for procedural task planning~\citep{chang2020procedure,bi2021procedure} which only plan for a fixed horizon of $T = 3$ or $T=4$ steps. We further visualize and analyze the plans generated on a few testing videos in Appendix~\ref{app:visualization}.

\begin{table}[!t]
\centering
\caption{Plan generation results}
\label{tab:full_plan_results}
\begin{tabular}{ccc}
\thickhline
Metrics & \makecell{Complete plan\\generation} & \makecell{Planning \\after a prefix}  \\ \hline
Accuracy & \makecell{0.190\\\small($\pm$0.010)}  & \makecell{0.294\\\small($\pm$0.018)} \\
\hline mIoU & \makecell{0.333 \\\small($\pm$ 0.062)}  & \makecell{0.628\\\small($\pm$ 0.023)} \\
\thickhline
\end{tabular}
\end{table}

\subsection{Ablations}

We conducted ablation studies to evaluate the effectiveness of different components in the ADTG pipeline.

\subsubsection{Role of Action Embeddings}
First, we investigate the effectiveness of learning action embeddings as (pre/post)-condition transformations. We consider three alternative ways to generate action embedding vectors: \emph{Random} embeddings where the vectors are randomly initialized, \emph{One-hot} embeddings which are one-hot action encodings, and \emph{NTG} embeddings which uses the intermediate action encodings from the node localizer of the NTG baseline. The evaluation results in Table~\ref{tab:embedding_ablation_results} show that the four variants do not differ significantly, but the ADTG embedding scheme still achieves close-to-highest performance in all tasks. This suggests that treating actions as transformations leads to (only marginally) better representations, however it does not play a dominant role on the CrossTask dataset. We believe that this is primarily because the actions in the CrossTask tasks do not have much semantic or hierarchical overlap across tasks, unlike in \citet{wang2016actions}. Hence, any embedding which disambiguates actions clearly performs well on this dataset. We defer investigation of this effect with connected action semantics to future work.
\addtolength{\tabcolsep}{-1.2pt}
\begin{table}[!t]
\centering
\caption{Action embedding ablation results}
\label{tab:embedding_ablation_results}
\begin{tabular}{ccccc}
\thickhline
Metrics & ADTG & Random & One-hot & NTG embed \\ \hline
\makecell{Task tracking\\ accuracy} & \makecell{0.741\\\small($\pm$0.013)}  & \makecell{0.721\\\small($\pm$0.010)} & \makecell{0.728\\\small($\pm$0.016)}  & \makecell{\textbf{0.7770}\\\small($\pm$0.010)} \\
\hline 
\makecell{Task tracking \\acc. excl. null} & \makecell{\textbf{0.557}\\\small($\pm$0.042)} & \makecell{0.521\\\small($\pm$0.010)}  & \makecell{0.508\\\small($\pm$0.031)} & \makecell{0.549\\\small($\pm$0.034)} \\
\hline 
\makecell{Next action \\accuracy} & \makecell{\textbf{0.523}\\\small($\pm$0.026)} & \makecell{0.459\\\small($\pm$0.031)} & \makecell{0.470\\\small($\pm$0.032)} & \makecell{0.500\\\small($\pm$0.017)} \\
\hline 
\makecell{Planning\\accuracy}  & \makecell{0.294\\\small($\pm$0.018)} & \makecell{0.279\\\small($\pm$0.022)} & \makecell{\textbf{0.311}\\\small($\pm$0.013)} & \makecell{0.306\\\small($\pm$0.014)} \\
\hline 
\makecell{Planning\\mIoU} & \makecell{0.628\\\small($\pm$0.023)} & \makecell{\textbf{0.631}\\\small($\pm$0.037)} & \makecell{0.596\\\small($\pm$0.030)} & \makecell{0.606\\\small($\pm$0.052)} \\ \thickhline
\end{tabular}
\end{table}
\addtolength{\tabcolsep}{1.2pt}

\subsubsection{Role of Action History}
In the second ablation, we evaluate whether using an action history in the task tracking and next action recommendation modules are helpful. We compare with a variant of ADTG that removes the history RNN and does not rely on the action history information. The results are presented in Table~\ref{tab:history_ablation_results}. We see that ADTG significantly outperforms its no-history variant on all evaluation metrics, thereby assuring that action history information is very helpful in procedural tasks.

\begin{table}[!h]
\centering
\caption{Action history ablation results}
\label{tab:history_ablation_results}
\begin{tabular}{ccc}
\thickhline
Metrics & ADTG & ADTG (no history) \\ \hline
Task tracking acc.  & \makecell{\textbf{0.741}\\\small($\pm$0.013)}  & \makecell{0.666\\\small($\pm$0.016)} \\
\hline \makecell{Task tracking\\ acc. excl. null} & \makecell{\textbf{0.557} \\\small($\pm$ 0.042)}  & \makecell{0.443\\\small($\pm$ 0.033)} \\
\hline Next action acc.  & \makecell{ \textbf{0.523} \\\small($\pm$ 0.026)} &  \makecell{0.413\\\small($\pm$ 0.026)}  \\
\hline Planning acc. & \makecell{\textbf{0.294} \\\small($\pm$0.018)} &  \makecell{0.204 \\\small($\pm$0.014)} \\
\hline Planning mIoU   & \makecell{\textbf{0.628} \\\small($\pm$ 0.023)} &  \makecell{0.376\\\small($\pm$ 0.033)}  \\ \thickhline
\end{tabular}
\end{table}

\section{Discussion}

From the experimental results and ablation studies, we have demonstrated that using the action history information contributes significantly to the superior performance of ADTG. This suggests that, for tasks where the ordering and temporal dependencies of the actions play a key role, one must condition on the action history. In addition, compared to existing end-to-end architectures, another advantage of ADTG is the modular design, which allows us to share certain modules across multiple downstream evaluation tasks, and to evaluate the performance of the pipeline in finer granularity.

A potential improvement would be to make better use of the visual observations in plan generation. The current design of the ADTG pipeline is not able to correct itself in the later steps of plan generation if it localizes itself in the wrong action node at the very beginning, which accounts for the failure cases in our visualized qualitative study (see Appendix~\ref{app:visualization}). Interesting future directions would include incorporating more state information (e.g., object or attribute information) on the edges of ADTG graphs, and modifying the beam search for generating plans to utilize this information without having to make explicit state predictions.

\section{Conclusion}
In this paper, we have presented an Action Dynamics Task Graphs approach for learning structured representations of procedural tasks from video demonstrations. We have shown that ADTG can be used to provide guidance to a human user via task tracking, next action recommendation, and plan generation. We have conducted experiments on the CrossTask dataset and demonstrated the superior performance of ADTG over existing baselines. 

\section*{Acknowledgments}

We sincerely thank Rohan Chitnis for valuable feedback and comments.

\bibliography{aaai23}

\begin{thebibliography}{19}
\providecommand{\natexlab}[1]{#1}

\bibitem[{Bi, Luo, and Xu(2021)}]{bi2021procedure}
Bi, J.; Luo, J.; and Xu, C. 2021.
\newblock Procedure planning in instructional videos via contextual modeling
  and model-based policy learning.
\newblock In \emph{Proceedings of the IEEE/CVF International Conference on
  Computer Vision}, 15611--15620.

\bibitem[{Carreira and Zisserman(2017)}]{carreira2017quo}
Carreira, J.; and Zisserman, A. 2017.
\newblock Quo vadis, action recognition? a new model and the kinetics dataset.
\newblock In \emph{Proceedings of the IEEE Conference on Computer Vision and
  Pattern Recognition}, 6299--6308.

\bibitem[{Chang et~al.(2020)Chang, Huang, Xu, Adeli, Fei-Fei, and
  Niebles}]{chang2020procedure}
Chang, C.-Y.; Huang, D.-A.; Xu, D.; Adeli, E.; Fei-Fei, L.; and Niebles, J.~C.
  2020.
\newblock Procedure planning in instructional videos.
\newblock In \emph{European Conference on Computer Vision}, 334--350. Springer.

\bibitem[{Fang et~al.(2020)Fang, Zhu, Garg, Savarese, and
  Fei-Fei}]{fang2020dynamics}
Fang, K.; Zhu, Y.; Garg, A.; Savarese, S.; and Fei-Fei, L. 2020.
\newblock Dynamics Learning with Cascaded Variational Inference for Multi-Step
  Manipulation.
\newblock In \emph{Conference on Robot Learning}, 42--52. PMLR.

\bibitem[{Hafner et~al.(2019)Hafner, Lillicrap, Fischer, Villegas, Ha, Lee, and
  Davidson}]{hafner2019learning}
Hafner, D.; Lillicrap, T.; Fischer, I.; Villegas, R.; Ha, D.; Lee, H.; and
  Davidson, J. 2019.
\newblock Learning latent dynamics for planning from pixels.
\newblock In \emph{International Conference on Machine Learning}, 2555--2565.
  PMLR.

\bibitem[{Hayes and Scassellati(2016)}]{hayes2016autonomously}
Hayes, B.; and Scassellati, B. 2016.
\newblock Autonomously constructing hierarchical task networks for planning and
  human-robot collaboration.
\newblock In \emph{IEEE International Conference on Robotics and Automation},
  5469--5476. IEEE.

\bibitem[{He et~al.(2016)He, Zhang, Ren, and Sun}]{he2016deep}
He, K.; Zhang, X.; Ren, S.; and Sun, J. 2016.
\newblock Deep residual learning for image recognition.
\newblock In \emph{IEEE Conference on Computer Vision and Pattern Recognition},
  770--778.

\bibitem[{Hershey et~al.(2017)Hershey, Chaudhuri, Ellis, Gemmeke, Jansen,
  Moore, Plakal, Platt, Saurous, Seybold et~al.}]{hershey2017cnn}
Hershey, S.; Chaudhuri, S.; Ellis, D.~P.; Gemmeke, J.~F.; Jansen, A.; Moore,
  R.~C.; Plakal, M.; Platt, D.; Saurous, R.~A.; Seybold, B.; et~al. 2017.
\newblock {CNN} architectures for large-scale audio classification.
\newblock In \emph{IEEE International Conference on Acoustics, Speech and
  Signal Processing}, 131--135.

\bibitem[{Huang et~al.(2019)Huang, Nair, Xu, Zhu, Garg, Fei-Fei, Savarese, and
  Niebles}]{huang2019neural}
Huang, D.-A.; Nair, S.; Xu, D.; Zhu, Y.; Garg, A.; Fei-Fei, L.; Savarese, S.;
  and Niebles, J.~C. 2019.
\newblock Neural task graphs: {G}eneralizing to unseen tasks from a single
  video demonstration.
\newblock In \emph{Proceedings of the IEEE/CVF Conference on Computer Vision
  and Pattern Recognition}, 8565--8574.

\bibitem[{Kingma and Ba(2014)}]{kingma2014adam}
Kingma, D.~P.; and Ba, J. 2014.
\newblock Adam: A method for stochastic optimization.
\newblock \emph{arXiv preprint arXiv:1412.6980}.

\bibitem[{Kurutach et~al.(2018)Kurutach, Tamar, Yang, Russell, and
  Abbeel}]{kurutach2018learning}
Kurutach, T.; Tamar, A.; Yang, G.; Russell, S.~J.; and Abbeel, P. 2018.
\newblock Learning plannable representations with causal {InfoGAN}.
\newblock \emph{Advances in Neural Information Processing Systems}, 31.

\bibitem[{Liu et~al.(2016)Liu, Yang, Saba-Sadiya, Shukla, He, Zhu, and
  Chai}]{liu2016jointly}
Liu, C.; Yang, S.; Saba-Sadiya, S.; Shukla, N.; He, Y.; Zhu, S.-C.; and Chai,
  J. 2016.
\newblock Jointly learning grounded task structures from language instruction
  and visual demonstration.
\newblock In \emph{Proceedings of the 2016 Conference on Empirical Methods in
  Natural Language Processing}, 1482--1492.

\bibitem[{Pan et~al.(2020)Pan, Chen, Wu, Liu, Ngo, Kan, Jiang, and
  Chua}]{pan2020multi}
Pan, L.-M.; Chen, J.; Wu, J.; Liu, S.; Ngo, C.-W.; Kan, M.-Y.; Jiang, Y.; and
  Chua, T.-S. 2020.
\newblock Multi-modal cooking workflow construction for food recipes.
\newblock In \emph{Proceedings of the 28th ACM International Conference on
  Multimedia}, 1132--1141.

\bibitem[{Srinivas et~al.(2018)Srinivas, Jabri, Abbeel, Levine, and
  Finn}]{srinivas2018universal}
Srinivas, A.; Jabri, A.; Abbeel, P.; Levine, S.; and Finn, C. 2018.
\newblock Universal planning networks: {L}earning generalizable representations
  for visuomotor control.
\newblock In \emph{International Conference on Machine Learning}, 4732--4741.
  PMLR.

\bibitem[{Sun et~al.(2022)Sun, Huang, Lu, Liu, Zhou, and Garg}]{sun2022plate}
Sun, J.; Huang, D.-A.; Lu, B.; Liu, Y.-H.; Zhou, B.; and Garg, A. 2022.
\newblock {PlaTe}: {V}isually-grounded planning with transformers in procedural
  tasks.
\newblock \emph{IEEE Robotics and Automation Letters}, 7(2): 4924--4930.

\bibitem[{Wang, Farhadi, and Gupta(2016)}]{wang2016actions}
Wang, X.; Farhadi, A.; and Gupta, A. 2016.
\newblock Actions\~{} transformations.
\newblock In \emph{Proceedings of the IEEE Conference on Computer Vision and
  Pattern Recognition}, 2658--2667.

\bibitem[{Xu et~al.(2020)Xu, Ji, Shi, Du, Neubig, Bisk, and
  Duan}]{xu2020benchmark}
Xu, F.~F.; Ji, L.; Shi, B.; Du, J.; Neubig, G.; Bisk, Y.; and Duan, N. 2020.
\newblock A benchmark for structured procedural knowledge extraction from
  cooking videos.
\newblock \emph{arXiv preprint arXiv:2005.00706}.

\bibitem[{Zhao et~al.(2022)Zhao, Hadji, Dvornik, Derpanis, Wildes, and
  Jepson}]{zhao2022p3iv}
Zhao, H.; Hadji, I.; Dvornik, N.; Derpanis, K.~G.; Wildes, R.~P.; and Jepson,
  A.~D. 2022.
\newblock {P3IV}: {P}robabilistic Procedure Planning from Instructional Videos
  with Weak Supervision.
\newblock In \emph{Proceedings of the IEEE/CVF Conference on Computer Vision
  and Pattern Recognition}, 2938--2948.

\bibitem[{Zhukov et~al.(2019)Zhukov, Alayrac, Cinbis, Fouhey, Laptev, and
  Sivic}]{zhukov2019cross}
Zhukov, D.; Alayrac, J.-B.; Cinbis, R.~G.; Fouhey, D.; Laptev, I.; and Sivic,
  J. 2019.
\newblock Cross-task weakly supervised learning from instructional videos.
\newblock In \emph{Proceedings of the IEEE/CVF Conference on Computer Vision
  and Pattern Recognition}, 3537--3545.

\end{thebibliography}



\appendix
\clearpage

\section{Details on the CrossTask Dataset}
\label{app:dataset}

In this appendix, we provide detailed statistics for the 18 primary tasks in the CrossTask dataset~\cite{zhukov2019cross}. This dataset contains 2750 videos, each demonstrating one of its 18 procedural tasks; e.g., ``Make a latte'', ``Change a tire'', or ``Make pancakes''. The average video length is about 5 minutes, with a total of 212 hours of recorded videos. These tasks are fairly complex and each task takes on average 7.4 actions to complete. Simpler tasks like ``Jack up a car'' take about 3 actions to finish, while more complicated ones like ``Change a tire'' can take as many as 11 actions.

Table~\ref{tab:crosstask_stats} shows a complete list of the 18 primary tasks, their average video lengths, sizes of the action spaces, average step lengths, and percentage of ``null action'' steps. The action space of a task is the set of all candidate actions that can be taken when performing the task, although some videos skip certain actions and do not cover the complete action space. The average step length measures the average number of steps taken in the actual video demonstrations to complete a task. Due to certain actions being skipped in some videos and others being repeated, the average step length of a task is not always equal to its action space size. ``Null actions'' are used to refer to the video segments that do not have actual actions happening, such as the introduction part of the video or transitioning scenes from one action to another. In Figure~\ref{fig:example-ctgs-app}, we also illustrate the Action Dynamics Task Graphs for several tasks from the CrossTask dataset.

For reproducibility and fair comparison with existing methods, we leverage the pre-computed video features provided along with the CrossTask dataset. For each one-second segment of the video, a 3200-dimensional feature vector is provided and contains a concatenation of 1024-D RGB I3D features~\citep{carreira2017quo}, 2048-D Resnet-152 features~\citep{he2016deep}, and 128-D audio VGG features~\citep{hershey2017cnn}.

\begin{table*}[!htbp]
\centering
\caption{Statistics of the CrossTask dataset. }
\label{tab:crosstask_stats}
\begin{tabular}{ccccc}
\thickhline
Task & Number of videos & Action space size & Average step length & Percentage of null action \\ \hline
Make Jello Shots & 182 & 6 & 7.90  & 72\% \\
Build Simple Floating Shelves & 153 & 5 & 5.54 &  58\%\\
Make Taco Salad & 170 &  8& 6.34 &  79\%\\
Grill Steak & 228 &  11 &  8.54 & 75\% \\
Make Kimchi Fried Rice & 120 & 6 & 8.66 & 70\% \\
Make Meringue & 154 & 6 & 6.72 &  67\%\\
Make a Latte & 157 & 6 & 5.06 &  71\%\\
Make Bread and Butter Pickles & 106 & 11 & 6.44 & 75\% \\
Make Lemonade & 131 & 8 & 8.28 &  69\%\\
Make French Toast & 252 & 10 & 9.10 &  68\%\\
Jack Up a Car &  89& 3 & 3.38 & 81\% \\
Make Kerala Fish Curry & 149 & 7 & 10.02 & 69\% \\
Make Banana Ice Cream & 170 & 5 & 4.52 & 80\% \\
Add Oil to Your Car & 137 & 8 & 8.04 &85\%  \\
Change a Tire & 99 & 11 & 9.84 & 62\% \\
Make Irish Coffee & 185 & 5 & 4.94 &  74\%\\
Make French Strawberry Cake &86  &  9& 11.56 &  63\%\\
Make Pancakes & 182 & 8 & 10.54 &  70\%\\ \hline
Average & 153 &7.4  & 7.84 & 72\% \\ \thickhline
\end{tabular}
\end{table*}

\section{Details for the CrossTask Baseline}
\label{app:crosstask}
The CrossTask baseline refers to the solution proposed in the same work~\citep{zhukov2019cross} along with the CrossTask dataset. The CrossTask baseline is a weakly-supervised approach for learning from instructional videos. It does not rely on the strong supervisions via temporal annotations of the action boundaries, but instead only use the temporal constraints generated from the instructional narrations and an ordered list of the action steps. The CrossTask approach is built upon the idea that the learning model should share certain components (e.g., verbs or nouns) while learning different steps across multiple tasks. For example, the action ``pour egg'' should be trained jointly with other tasks involving the components ``pour'' or ``egg''. Following this idea, CrossTask proposes to use component models to represent each step as its constituent components instead of as a monolithic entity. The step assignment objective in CrossTask essentially corresponds to our task tracking module, yet CrossTask does not support next action recommendation or plan generation. 

Since the original CrossTask approach is a weakly-supervised learning method, for fair comparisons, in our experiments we also consider a supervised-learning variant of CrossTask that adopts the same linear classifier as~\citep{zhukov2019cross}, but further uses the annotated action segmentation boundaries for training. 

\begin{figure*}[!htbp]
\centering
\includegraphics[width=.8\textwidth]{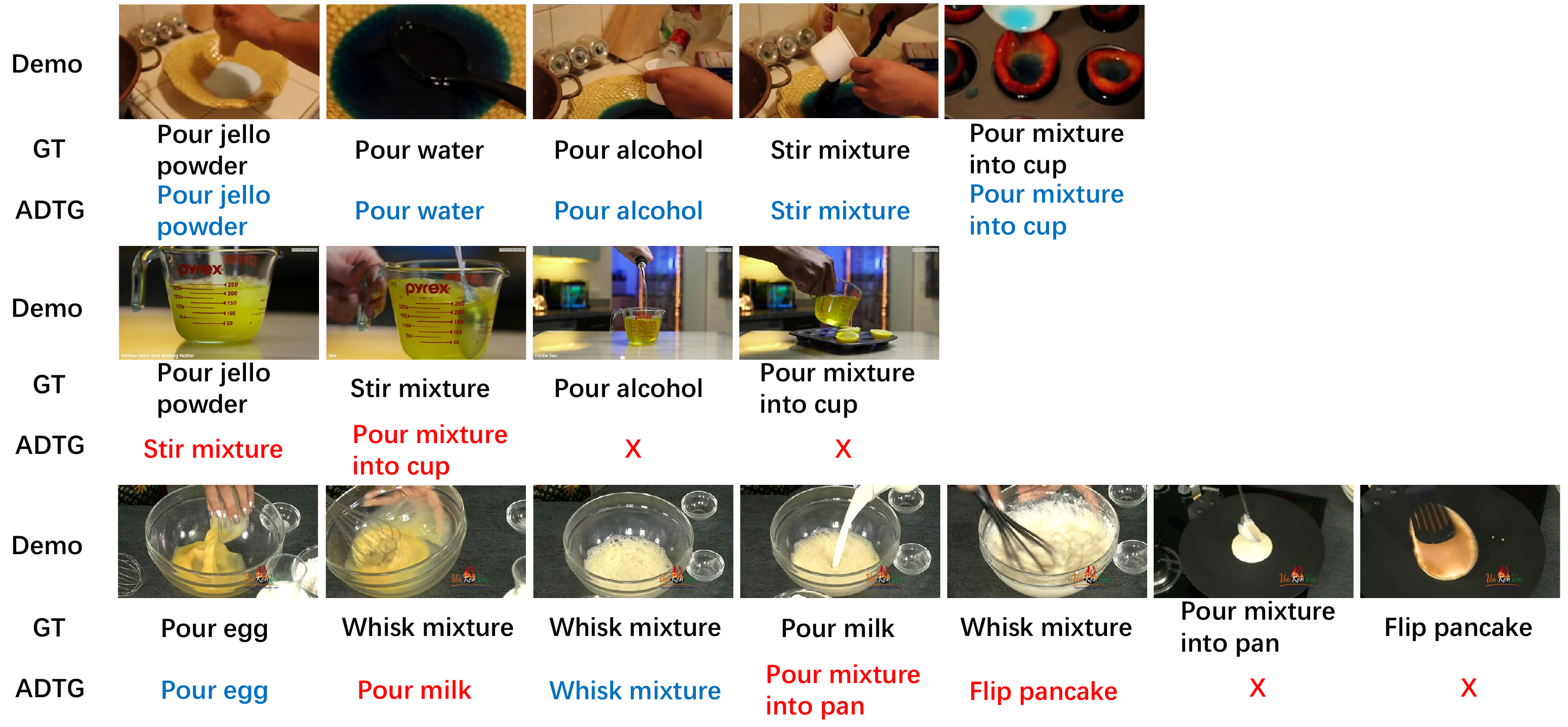}
\caption{Visualization of ground-truth plans (GT) vs. ADTG generated plans. }
\label{fig:plan_viz}
\end{figure*}

\section{Details for the Neural Task Graphs Baseline}
\label{app:ntg}
Similar to ours, the NTG approach is a modularized method that uses (a simplified variant of) the conjugate task graphs as intermediate representations. NTG focuses on generalizing to unseen tasks from a single video demonstration in the same domain. It uses the CTG representations to explicitly modularize the video demonstration and the derived policy, so as to incorporate the compositional structure of the tasks into the NTG model. Specifically, NTG consists of a generator that builds a conjugate task graph from video demonstrations, and an execution engine that uses the learned tasks graphs to perform task tracking. In particular, the NTG generator itself can be decomposed into two parts: a demo interpreter that is used to obtain a single action path traversing the CTG by observing the action sequence in the video demonstration, and a graph completion network that adds the edges that are not observed in the single demonstration to capture the potential interchangeability of the action ordering. The NTG execution engine also consists of two parts: A node localizer that tries to localize the current action node in the CTG based on the visual observation (i.e., task tracking), and an edge classifier that checks the  precondition of each possible outgoing edge from the localized node to decide the next action (i.e., next action recommendation). Since the edge classifier in NTG relies on visual observations as input, in the absence of an interactive environment, it cannot generate a full plan in an autoregressive way as we do.

\section{Plan Visualization}\label{app:visualization}
In Figure~\ref{fig:plan_viz}, we visualize the planned action sequences generated by ADTG on a few testing videos, and compare them with the ground-truth (GT) plans. In the first example, ADTG successfully generates the correct sequence of actions for the task ``Make jello shots''. In the second example, the task tracking module of ADTG fails to recognize the first step (``pour jello powder'') of the video and misclassifies it as ``stir mixture''. Since ADTG generates plans by recursively invoking the next action recommendation module, it is not able to correct such a mistake and hence diverges from the ground-truth action sequence afterward. In the last example (on the task ``Make pancakes''), even though the action sequence planned by ADTG does not exactly match the ground-truth plan, it still forms a semantically reasonable plan to complete the task. This is because the ADTG generated plan simply switches the order of the actions ``pour milk'' and ``whisk mixture'' compared to the ground-truth and removes the repeated ``whisk mixture'' steps, which makes sense in the given task. This also suggests that we might need better ways to evaluate plans in such datasets that do not have an interactive environment and we leave this to future work.

\end{document}